\documentclass{ieeeaccess}
\usepackage{cite}
\usepackage{amsmath,amssymb,amsfonts}
\usepackage{algorithmic}
\usepackage{graphicx}
\usepackage{textcomp}
\usepackage{caption}

\def\BibTeX{{\rm B\kern-.05em{\sc i\kern-.025em b}\kern-.08em
    T\kern-.1667em\lower.7ex\hbox{E}\kern-.125emX}}
\begin{document}
\history{Date of publication xxxx 00, 0000, date of current version xxxx 00, 0000.}
\doi{10.1109/ACCESS.2017.DOI}

\title{On the Initial Behavior Monitoring Issues in Federated Learning}
\author{\uppercase{Ranwa Al Mallah}\authorrefmark{1}, \IEEEmembership{Member, IEEE},
\uppercase{Godwin Badu-Marfo\authorrefmark{2}, and Bilal Farooq}.\authorrefmark{3},
\IEEEmembership{Member, IEEE}}
\address[1]{Department of Electrical and Computer Engineering, Royal Military College of Canada, Kingston, ON, Canada  (e-mail: almallah@rmc-cmr.ca)}
\address[2]{Department of Civil Engineering, Ryerson University, Toronto, 
ON, Canada (e-mail: gbmarfo@ryerson.ca)}
\address[3]{Department of Civil Engineering, Ryerson University, Toronto, 
ON, Canada (e-mail: bilal.farooq@ryerson.ca)}

\tfootnote{This research is partially funded by Canada Research Chair program.}

\markboth
{Author \headeretal: Preparation of Papers for IEEE TRANSACTIONS and JOURNALS}
{Author \headeretal: Preparation of Papers for IEEE TRANSACTIONS and JOURNALS}

\corresp{Corresponding author: Bilal Farooq (e-mail: bilal.farooq@ryerson.ca).}

\begin{abstract}
In Federated Learning (FL), a group of \textit{workers} participate to build a global model under the coordination of one node, the \textit{chief}. Regarding the cybersecurity of FL, some attacks aim at injecting the fabricated local model updates into the system. Some defenses are based on malicious \textit{worker} detection and behavioral pattern analysis. In this context, without timely and dynamic monitoring methods, the \textit{chief} cannot detect and remove the malicious or unreliable \textit{workers} from the system. Our work emphasizes the urgency to prepare the federated learning process for monitoring and eventually behavioral pattern analysis. We study the information inside the learning process in the early stages of training, propose a monitoring process and evaluate the monitoring period required. The aim is to analyze at what time is it appropriate to start the detection algorithm in order to remove the malicious or unreliable \textit{workers} from the system and optimize the defense mechanism deployment. We tested our strategy on a behavioral pattern analysis defense applied to the FL process of different benchmark systems for text and image classification. Our results show that the monitoring process lowers false positives and false negatives and consequently increases system efficiency by enabling the distributed learning system to achieve better performance in the early stage of training. 
\end{abstract}


\titlepgskip=-15pt

\maketitle

\section{Introduction}
\label{sec:introduction}
\PARstart{F}{ederated} Learning (FL) is transforming many industries including defense, telecommunications, IoT, and transportation \cite{yang2019federated}. This Machine Learning (ML) paradigm addresses issues such as data privacy, security, access rights and access to heterogeneous data by training a global model using distributed data. In FL, a group of nodes called \textit{workers} participate to build a global model under the coordination of one node, the \textit{chief}. After each round of the FL process, the \textit{chief} sends a matrix of weights to the \textit{workers}. Each \textit{worker} then computes its local model update based on their local data. Once the \textit{chief} node gets all the local model updates from the \textit{workers}, it averages the models with an aggregation rule and updates them into a single matrix. The process repeats for several rounds of training until a pre-specified test accuracy is reached or the maximum number of time epochs have elapsed.

Although FL enables nodes to construct a model without sharing their private data with others, the fact that the \textit{chief} has no visibility into how the updates are generated represents a vulnerability in the process. Regarding the cybersecurity of FL, attacks aim at exploiting this vulnerability. According to Blanchard et al. \cite{blanchard2017machine}, classical averaging rules are so fragile that even a single malicious \textit{worker} can manipulate the entire learning process by sabotaging prediction accuracy. 
Many defense mechanisms propose corrections to the FL process to increase system efficiency. They apply statistical methods to correct the aggregation rule implemented at the \textit{chief}  \cite{chen2017distributed, mhamdi2018hidden, alistarh2018byzantine}. Rather than correcting, other approaches proposed to detect malicious \textit{workers} by interpreting information of the \textit{workers}’ behaviors in order to eliminate potential malicious \textit{workers}. Kang et al. proposed a reputation-based scheme in order to select reliable and trusted \textit{workers} \cite{kang2020reliable} . Preuveneers et al.  proposed to audit the local model updates and the \textit{workers} in the FL process are held accountable \cite{preuveneers2018chained}. 

As most of the defense strategies are based on malicious \textit{worker} detection, there is an urgency to prepare the federated learning process for monitoring and eventually behavioral pattern analysis. In fact, without timely and dynamic monitoring methods, the \textit{chief} cannot detect and remove the malicious or unreliable \textit{workers} from the system. Moreover, it is hard for the \textit{chief} to monitor the large-scale \textit{worker} behaviors in real-time. Pan et al. scratched the surface by proposing to visualize the behavior of the \textit{workers} \cite{pan2020justinian}. They suggested to visualize the policy sequence of their model in order to understand the temporal patterns of the attacks.

This paper proposes to study the information inside the learning process of each \textit{worker} in order to increase the efficiency of the behavior pattern analysis approaches. Precisely, we aim to analyse at what time is it appropriate to start the detection algorithm in order to remove the malicious or unreliable \textit{workers} from the system. The intuition is that \textit{workers} should not be removed from the FL process too early nor too late. There must be an appropriate time window for monitoring after which the detection protocol or the behavior pattern analysis will start declaring that the nodes are malicious. Otherwise, in the early stages of training, this will result in false evaluation of the legitimacy of the nodes.  

Moreover, in a recent review of the topic, Lyu et al. \cite{lyu2020privacy} highlighted the importance of studying how to optimize the timing of deploying the defense mechanisms. They stressed the fact that when defense mechanisms are deployed to check if any adversary is attacking the FL system, the FL server requires additional computational cost. They also put forward that different types of defense mechanisms may incur different cost against different attacks. Our paper aims at addressing the challenges by proving that there must be a monitoring period during the FL process prior to the deployment of the defenses. In our analysis, we show that the time metric has an impact on the false positive and false negative rates. By studying the context of the federated learning process, the defense mechanisms must compute the time metric as per our analysis in order to provide more accurate and efficient results. We tested our suggested monitoring process and empirical results show a decrease in false positive and false negative rates and consequently, an increase in system efficiency. The key contributions of this paper are:

\begin{itemize}
    \item A monitoring process during the early stages of training in FL for the defense approaches that are based on behavioral pattern analysis to eliminate potential malicious \textit{workers} from the FL process. 
    \item Evaluation of the monitoring period required in order to improve system efficiency in terms of the time cost of the iterations and computation overheads in the beginning of training for the behavioral pattern analysis approaches.
    \item Implementation of the monitoring process and evaluation of its appropriate time period on a behavioral pattern analysis defense applied to the distributed learning process of different benchmark systems for text and image classification (MNIST, CIFAR-10). 
\end{itemize}

The rest of the paper is organised as follows: in Section~\ref{sec:background} we present a literature review on the cybersecurity of the FL process. In Section~\ref{sec:methodology} is described the methodology. In Sections ~\ref{sec:implementation} and ~\ref{sec:Results}, we describe the implementation details, the simulation and provide experimental results of the monitoring process. The primary conclusions and future work are outlined in Section~\ref{sec:conclusion}.

\section{Literature review}
\label{sec:background}

Typically, the FL process consists of $\mathcal{K}$ \textit{workers} and a \textit{chief} that trains a model \textit{f} with $\textbf{w}_G$ $\in$ $\mathbb{R}^n$ being the global parameter vector, where \textit{n} is the dimension of the parameter space. The resulting global model is obtained by distributed training and aggregation over the $\mathcal{K}$ \textit{workers} holding data samples $\mathcal{D}_i$ locally, where $\mid\mathcal{D}_i\mid = l_i $. The total number of data samples is $ \sum\limits_{i} l_i = l$. Each \textit{worker} keeps its data private, i.e. $\mathcal{D}_i$ = $\{x_1^i, ..., x^i_l\}$ is not shared with the \textit{chief} $\mathcal{S}$ or any other \textit{worker} $i \neq j, \forall j \in \mathcal{K}$ \cite{yang2019federated}. 

The aim of the training is to generalize beyond a test dataset $\mathcal{D}_{test}$. At each time step \textit{t}, a random subset of \textit{k} \textit{workers} is selected by the \textit{chief}. Every \textit{worker} $ i \in \mid k \mid$ trains a local model to minimize a loss function over its own data $\mathcal{D}_i$. The \textit{worker} starts the local training from the global model $\textbf{w}_G^t$ received from the \textit{chief} and runs an algorithm such as the stochastic gradient descent for a number of $\mathcal{E}$ epochs with a batch size of $\mathcal{B}$. Each \textit{worker} then obtains a local weight vector $\textbf{w}_i$ and then computes its local model update $\delta_i^t$ = $\textbf{w}_i$ - $\textbf{w}_G^t$. The local model update is sent back to the \textit{chief}. To obtain the global model update $\textbf{w}_G$ for the next iteration, an aggregation mechanism is used: $\textbf{w}_G$ = $\textbf{w}_G^t$ + $ \sum\limits_{ i \in \mid k \mid} \alpha_i \delta_i$, where $\alpha_i$ = $l_i/l$ and $ \sum\limits_{i} \alpha_i = 1$.

Regarding the cybersecurity of the FL process, some statistical methods aim at correcting the aggregation rule implemented at the chief node. GeoMed computes the geometric median as the proposed estimator, which assumes the Byzantine ratio satisfies n $\geqslant$ 2m + 1 \cite{chen2017distributed}. 
Krum is another approach that assumes the Byzantine ratio satisfies n $\geqslant$ 2m + 3 \cite{blanchard2017machine}. It first finds the $n - m - 2$ closest vectors in Q for each $V_i$ , and then computes a score for each vector $V_i$. Finally, it proposes the vector $V_i$ with the smallest score as the next update step, i.e. $F(V_1, ..., V_n) = argmin_Vi\in_Q s(V_i)$. Bulyan was originally designed for Byzantine attacks that concentrate on a single coordinate \cite{mhamdi2018hidden}. First, it runs Krum over \textit{Q} without replacement for $n-2m$ time and collect the $n - 2m$ gradients to form a selection set. It then computes \textit{F} coordinate-wise: the i-th coordinate of \textit{F} is equal to the average of the $n-4m$ closest i-th coordinates to the median i-th coordinate of the selection set. Bulyan has the strictest assumption as n $\geqslant$ 4m+3 which significantly limits its practical usage.

In \cite{fang2020local}, to defend against local model poisoning attacks, the authors generalize RONI \cite{barreno2010security} and TRIM \cite{jagielski2018manipulating}. Both generalized defenses remove the local models that are potentially malicious before computing the global model in each iteration of federated learning and they start as soon as training starts. One generalized defense removes the local models that have large negative impact on the error rate of the global model (inspired by RONI that removes training examples that have large negative impact on the error rate of the model), while the other defense removes the local models that result in large loss (inspired by TRIM that removes the training examples that have large negative impact on the loss). They do not consider delaying the anomaly detection by leaving a monitoring period before starting the defense. 

In \cite{cao2019understanding}, the authors propose a filtering mechanism, Sniper, which is conducted by the chief, to remove attackers from the global model aggregation by examining Euclidean distances between local models. Through observing the distance between models, they found models of honest users and attackers are in different cliques. Based on this, they propose a filtering defense mechanism where during every communication, the chief must run it to filter parameters updated by attackers. As a result, Sniper can recognize honest users and drop attack success rate significantly even when multiple attackers are in the federated learning system. The chief runs Sniper during every updating rather than only at the beginning of training. This can prevent attackers upload contaminated parameters at a few epochs or after a few benign sharing. Again, they do not account for a monitoring period before starting the anomaly detection. We argue that monitoring period results in a better detection performance.

In \cite{fung2018mitigating}, they evaluate the vulnerability of federated learning to sybil-based poisoning attacks. They propose FoolsGold, a defense that identifies poisoning sybils based on the diversity of client updates in the distributed learning process. FoolsGold changes the federated learning algorithm, relies on standard techniques such as cosine similarity, and does not require prior knowledge of the expected number of sybils. FoolsGold uses client contribution similarity and starts the defense without leaving a monitoring period. 

An increasing number of approaches propose to detect malicious workers by interpreting information of the workers’ behaviors in order to eliminate them from the FL process. By eliminating potential malicious workers, the techniques constitute further mitigation and help accelerate the learning process. Pan et al. propose a reinforcement learning method based on a gradient aggregation agent \cite{pan2020justinian}. The method utilize the historical interactions with the workers as experience to generate reward signals for policy learning. They start the detection which they defined as a classification task with no monitoring period and report their results after a fixed number of iterations of distributed learning. Xie et al. use the loss decrease at the current iteration on the training set to rank the workers’ credibility \cite{xie2019zeno}. Afterwards, the algorithm aggregates the candidates with highest scores. The score roughly indicates the level of trust in each candidate. They start their suspicion-based aggregation rule without a monitoring period and eliminate malicious workers from the process by taking the gradient estimators with the highest scores only.

We want to better analyse the convergence of the FL process with regards to time when the defenses are applied in order to account for more effective and accurate malicious \textit{worker} detection and behavior pattern analysis. To the best of our knowledge, no previous work tried to leave a monitoring period in order to evaluate if it results in a better detection performance.

\section{Methodology}
\label{sec:methodology}

Defense mechanisms that are implemented at the \textit{chief} node try to learn from the interactions with the \textit{workers} and leverage the information received to protect the FL process in adversarial settings. The defense restores robustness by detecting malicious \textit{workers} and eliminating them from the FL process. In the current state of information technology security practice, detection and response come only after effective monitoring. This will allow to adequately identify malicious \textit{workers} and take appropriate response actions.

\subsection{Security settings}

\begin{figure}[h]
\centering
\includegraphics[width=1\columnwidth]{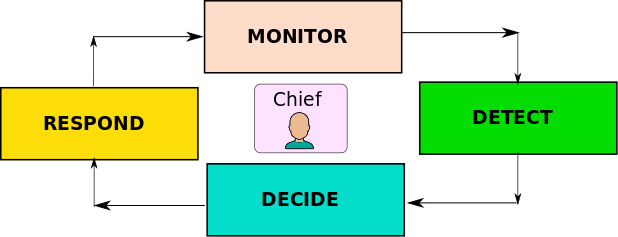}
\caption{Security decision loop - The cycle of security defense can be described in terms of a loop with the functions of Monitor, Detect, Decide and Respond.}
\label{fig:cycle}
\end{figure}

Figure~\ref{fig:cycle} shows the security decision loop enabling adequate decision-making in terms of detection, prioritization and response to attacks. We stress the fact that to maintain an adequate security posture, the monitoring period is required especially in FL where early detection will necessary get a significant number of false positives and false negatives in the beginning of training. Monitoring has to be done for a certain period of time after which the cycle continues with the detect, decide and respond actions as per the figure.

This work aims at studying the appropriate initial monitoring period of time $\Delta$ to consider in each context of FL. We study how long do we need to monitor before starting to detect. Instead of a random value of $\Delta$ after which evaluating the performance of the defense methods, we put forward a strategy to compute this value. If we are able to estimate the value of $\Delta$, we will improve the efficiency of FL in adversarial settings especially in terms of time efficiency where the running time of different defenses will improve. Precisely, the time cost of the iterations in the beginning of training will be greatly improved because only monitoring will be done during a period of $\Delta$. After this period, the defense algorithm may start to detect malicious workers and this would have brought computation overheads to lower levels. This approach is even more appreciated in a decentralized federated learning setting or in IoT networks, such as wearable devices, or smart homes where the edge device has limited resources or connectivity. In these settings, federated learning methods must be optimized to operate and train models that efficiently adapt to changes in these systems. Improving the efficiency of FL in adversarial settings especially in terms of time efficiency and computation overhead contributes to this achievement.

The value of the initial monitoring period of time $\Delta$ varies depending on many factors: the federated learning process under study (underlying case study for the training of the machine learning model), attacking patterns and defense strategy deployed. In fact, different FL processes for training deep neural networks have learning curves that follow different distributions. Moreover, different attack patterns tend to cause instability in the learning curves of the FL process. Finally, depending on the defense strategy deployed, some may perform better than others that tend to yield accuracy that distribute in a slightly wider range. 

\subsubsection{Attack patterns} 
In our threat model, malicious \textit{workers} create malicious local model updates by submitting random values sampled from a multi-dimensional Gaussian $\mathcal{N}(\mu,\,\sigma^{2})$. We take $\mu$ = [0.5,...,0.5] $\in$ $R^d$ and $\sigma$ = $2x10^6$. This attack represents an untargeted poisoning on the FL process. Moreover, we investigate different attacking patterns. Firstly, in a static attack, a fixed number of \textit{workers} are compromised and perform continuous attacks during the entire learning process. Another attack pattern consists of a fixed number of \textit{workers} pretending to be benign in the first fixed number of rounds and start the attack afterwards. We call this attack the pretence attack. In a randomized attack, each compromised \textit{worker} is assigned with its role by the adversary. During the learning process, the \textit{worker} changes its role with a certain probability.

\subsubsection{Defense strategy}

We implement one of the defense strategies proposed in Mallah et al. \cite{mallah2021untargeted} for untargeted poisoning attack detection in federated learning via behavior attestation. The strategy assumes that the \textit{chief} node has a small validation dataset. Then, the \textit{chief} tests how the local models sent by the \textit{worker} predict on the same validation dataset iteration after iteration in order to assess if the \textit{worker} is reliable.  The \textit{worker}'s performance in terms of error rate compared to their own previous performance on that validation dataset will highlight if the node appears to be learning. 
The strategy removes the local models that have large negative impact on the error rate. The error is expected to be smaller over time when the node is not malicious. However, if the error doesn't improve nor decrease over time, then the node is considered unreliable. At iteration $t$, for each \textit{worker} $j$, we compute the error, $E_j^{t+1}$ of the current local model $LM_j^{t+1}$ on the validation dataset. We then compare it to the error at the previous iteration, $E_j^{t}$ as follows:

\begin{equation}
E_j^{t+1} - E_j^{t}= E_j^{t+1}(LM_j^{t+1}) - E_j^{t}(LM_j^{t})
\end{equation}

\subsection{Analytical formulation}

In this section, we will provide a mathematical formulation of the monitoring period. In fact, the value of the monitoring period of time $\Delta_T$ varies depending on many factors. It may vary depending on the federated learning process under study, the attack patterns and the defense strategies.

\begin{equation}
\Delta_T = \min(\alpha_F, \beta_A, \gamma_D)
\end{equation}

\begin{align*}   
\alpha_F &= \lfloor f(s, m, a) \rfloor\\
\beta_A &= \lfloor g(A_i) \rfloor\\
\gamma_D &= \lfloor h(D_j) \rfloor
\end{align*}

$\alpha_F$ is a function of the federated learning process under study \textit{f(s, m, a)}, the underlying case study for the training of the machine learning model because different FL processes for training deep neural networks have learning curves that follow different distributions. The FL process is characterized by the size of the training set \textit{s}, the type of model \textit{m}, (supervised, semi-supervised or unsupervised), and the architecture \textit{a}, (convolutional neural networks, auto-encoders, recurrent neural networks). $\beta_A$ is a function of \textit{$g(A_i)$} and corresponds to the different attack patterns (\textit{i}= static, pretence, randomized attacks), because different patterns tend to cause instability in the learning curves of the FL process. $\gamma_D$ is a function of \textit{$h(D_j)$} and corresponds to the defense strategy deployed. Some defenses may perform better than others that tend to yield accuracy that distribute in a slightly wider range (\textit{j}= KRUM, GeoMed, attestedFL). 

The monitoring period $\Delta_T$ corresponds to the minimum number of iterations required before the performance accuracy of the model diverges from the global optima. The minimum value corresponds to the minimum of the $\alpha_F$, $\beta_A$, $\gamma_D$ metrics. Training of a neural network stops when the error, i.e., the difference between the desired output and the expected output is below some threshold value or the number of iterations or epochs is above some threshold value. Thus, the value of $\Delta_T$ ranges from 0 to the threshold value implemented. $\Delta_T$ is equal to $\alpha_F$ if the attack is benign and the defense is robust. This means that if an attack has a small impact on the FL process and the defense was able to neutralize it very quickly, the monitoring period required would be in line with the baseline behavior. $\Delta_T$ will be equal to $\beta_A$ if the attack is more powerful than the defense. Thus, $\beta_A$ becomes the limiting factor and the monitoring period of time advised should enable the defense to start eliminating earlier the malicious nodes from the aggregation. Finally, depending on the type of defense, $\Delta_T$ can be equal to $\gamma_D$, thus enabling a larger monitoring period before any anomaly detection can take effect. 

We developed a search process that will provide an empirical validation of the analytical results. Different datasets are used to demonstrate experimentally, under different scenarios, the monitoring period in terms of delay.

\section{Implementation details}
\label{sec:implementation}

We implement two use cases of federated learning with Torch, a framework for deep learning \cite{collobert2002torch}. The first use case consist in training a fully connected neural network for the hand-written digital classification task on the MNIST dataset, with 10 workers. This public dataset contains 60,000 28×28 images of 10 digits for training and 10000 for testing. The model consists of 784 inputs, 10 outputs with soft-max activation and one hidden layer with 30 ReLu units. The dimension of parameters is 25,450. 

The second use case is training a ResNet-18 model for the image classification on the CIFAR-10 dataset with 10 workers. This dataset contains 60,000 28×28×3 images of 10 classes of objects for training and 10000 for testing. The standard model ResNet-18 has 18 end-to-end layers and 11173962 learnable parameters in total. In all use cases, each \textit{worker} shares a copy of the training set. We simulate the federated learning setting by sequential computation of gradients on randomly sampled mini-batches.

Although any other dataset could have been chosen to validate our approach, MNIST and CIFAR-10 are used as a worldwide machine learning benchmark. Because MNIST is a labeled dataset that pairs images of hand-written numerals with the name of the respective numeral, it can be used in supervised learning to train classifiers. CIFAR-10 is an established computer-vision dataset used for object recognition. Both datasets are publicly available and are the most widely used datasets for machine learning research. Their simplicity and ease of use are what make them so widely used and deeply understood. 

To evaluate the performance of the FL process during the first iterations of training, we need to conduct experiments where the FL process is under normal conditions (i.e. under no attack) and report the accuracy the global model attains iteration after iteration. We will do so for the two use cases under study. Also, we need to conduct experiments where each FL process is under different attacking patterns (i.e static, pretence and randomized attacks). Finally, we conduct experiments when the defense is applied. In this case, we will also report the precision and recall of the detection mechanisms during the first iterations of the learning process and that, for different values of $\Delta$.

\subsection{Analysis process} 

We present a process to investigate all possible combination of experiments to ensure a complete understanding of the FL performance during the first iterations of the learning process. The process is based on a One at A Time (OAT) analysis of the different factors affecting the accuracy of the model.

In fact, the value of the initial monitoring period of time $\Delta$ depends on three factors: the federated learning process under study, attacking patterns and defense strategy deployed. In a OAT analysis, the effect of each factor on the accuracy is investigated. To do so, factors are investigated separately and one at a time analysis of each factor have to be conducted by varying one factor at a time and keeping all other factors fixed. Thus we first conduct the experiments where the FL is under no attacks. The performance of FL in each use case will serve as a baseline for the comparison with the other experiments. We then do the experiments where there are attacks and no defenses. They serve as the upper bound for the evaluation of $\Delta$. Afterwards we apply the defense on the three different attacks at three different values of $\Delta$. The OAT analysis will result in a total of 26 experiments (i.e. 18 + 2 + 6 = 26). For the experiments when defenses are applied, we present the precision and recall of the methods at each iteration. Because we are experimenting with three different values of $\Delta$, precision and recall will help estimate the appropriate value of $\Delta$ to be considered for each use case.


\section{Experimental Results} 	
\label{sec:Results}

The performance of malicious \textit{workers} detection methods may be reported in terms of precision and recall as for any pattern recognition, information retrieval and classification in machine learning methods. Precision, which represents the positive predictive value, is the fraction of relevant instances among the retrieved instances. Recall, which represents the sensitivity, is the fraction of the total amount of relevant instances that were actually retrieved. 

\begin{figure}
\begin{center}
  \includegraphics[width=1\linewidth]{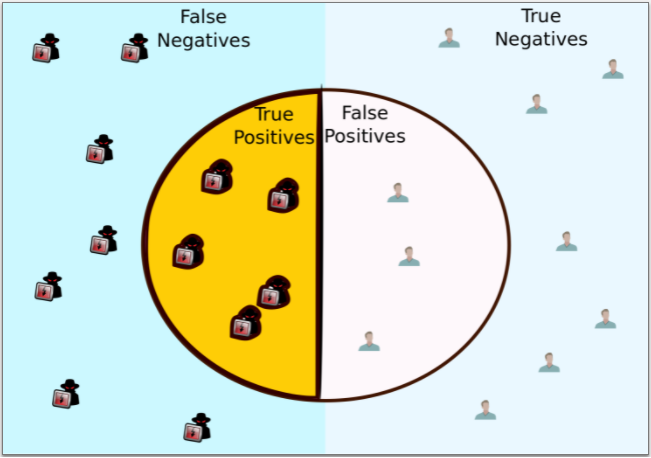}
  \caption{Components of precision and recall.}
  \label{fig:pr}
\end{center}
\end{figure}

In Figure~\ref{fig:pr}, we see a detection algorithm identifying 8 malicious \textit{workers} in an FL process containing 10 benign and 12 malicious \textit{workers} (the relevant elements). Of the 8 identified malicious \textit{workers}, 5 actually are malicious (True Positives, TP), while the other 3 are benign (False Positives, FP). 7 malicious \textit{workers} were missed (False Negatives, FN), and 7 benign were correctly excluded (True Negatives, TN). The detection’s precision is 5/8 (true positives / all positives) and represents how valid the results are. While its recall is 5/12 (true positives / relevant elements) and represents how complete the results are. 

In the context of FL, a detection technique that ensures high recall but reduces precision is better than one that increases precision but reduces recall. Greater recall increases the chances of removing benign \textit{workers} and increases the chances of removing all malicious workers. Greater precision decreases the chances of removing benign workers but also decreases the chances of removing all malicious \textit{workers}. In sum, it is not that bad if you remove some benign \textit{workers} from the FL process however it is bad if you don’t remove almost all malicious \textit{workers} because of the aggregation rule.

Among the strategies to evaluate the performance of the detection techniques, accuracy can be a misleading metric in the context of low ratio of malicious \textit{workers}. 





On the other hand, the F-measure considers precision and recall scores. In general, it is the harmonic mean, which, for the case of two numbers, coincides with the square of the geometric mean divided by the arithmetic mean. Particularly, the $F_1$ measure corresponds to a recall and precision that are evenly weighted and represents a special case of $\beta$ =1 in the general $F_\beta$ measure (for non-negative real values of $\beta$):

\begin{equation}
F_\beta = (1+ \beta^2) \bullet \frac{precision\bullet recall}{\beta^2 \bullet precision +recall}
\end{equation}

$F_\beta$  corresponds to the effectiveness of defense strategy with respect to a \textit{chief} who attaches $\beta$ times more importance to recall than precision". In our study, we use $\beta$=2 where the $F_2$ measure weights recall higher than precision. 
 
\subsection{Empirical validation}

\begin{figure}[!h]
\begin{center}
  \includegraphics[width=1\linewidth]{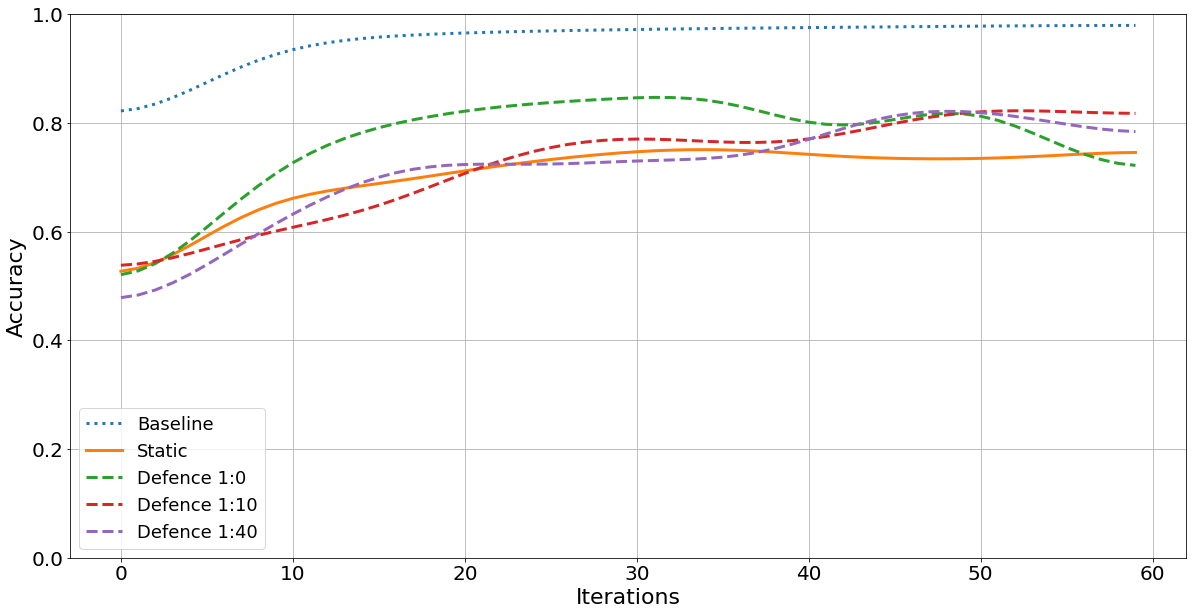}
  \caption{MNIST - Accuracy of the global model when under a static attack with the defense starting at different $\Delta$ periods.}
  \label{fig:MNIST_1}
\end{center}
\end{figure}

\begin{figure}[!h]
\begin{center}
  \includegraphics[width=1\linewidth]{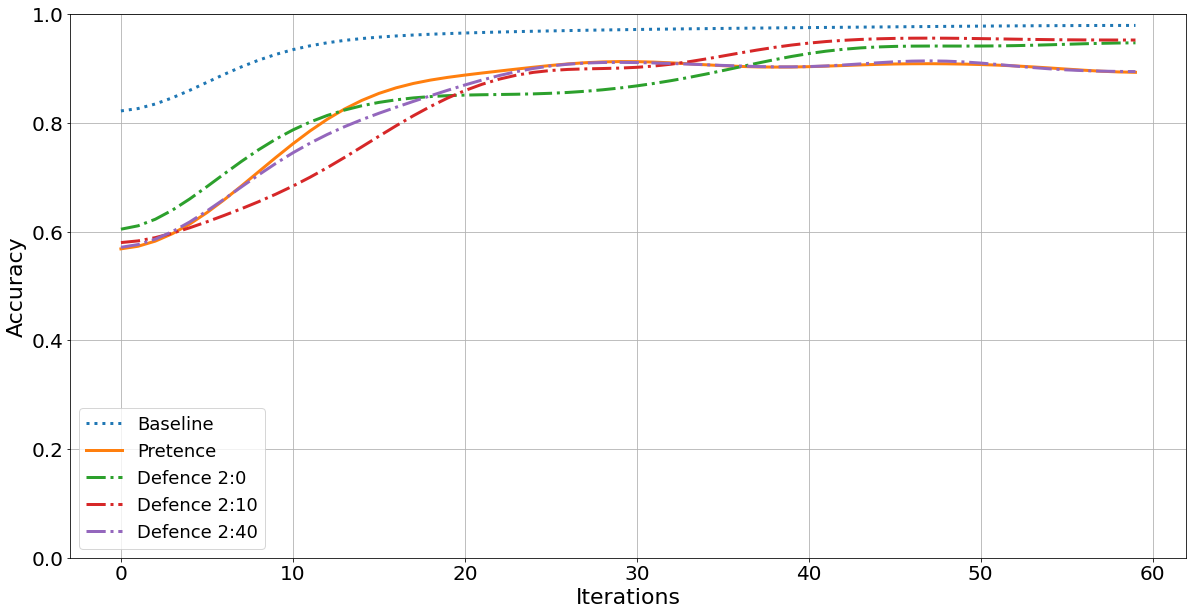}
  \caption{MNIST - Accuracy of the global model when under a pretence attack with the defense starting at different $\Delta$ periods.}
  \label{fig:MNIST_2}
\end{center}
\end{figure}

\begin{figure}[!h]
\begin{center}
  \includegraphics[width=1\linewidth]{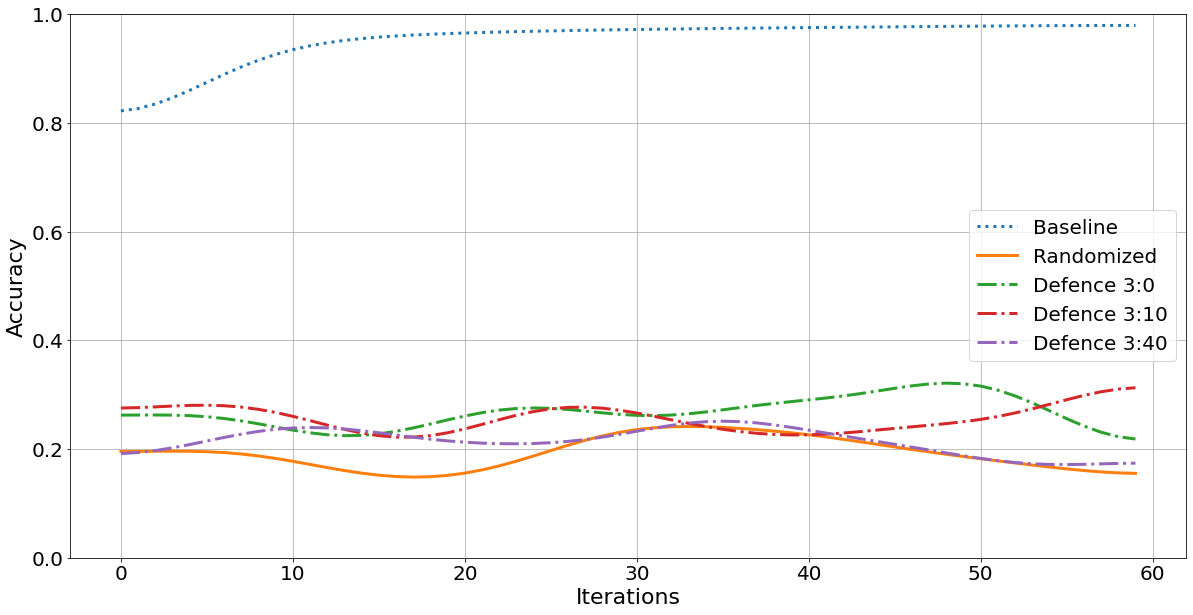}
  \caption{MNIST - Accuracy of the global model when under a randomized attack with the defense starting at different $\Delta$ periods.}
  \label{fig:MNIST_3}
\end{center}
\end{figure}

We run the two use cases (MNIST, CIFAR-10) with the defense starting at different monitoring periods of time $\Delta$ under different attacks. We show in Figures~\ref{fig:MNIST_1}, ~\ref{fig:MNIST_2}, and  ~\ref{fig:MNIST_3}, the accuracy of the global model during the MNIST FL process under different attacks. 

We notice from the figures that under a static attack on MNIST, the model's accuracy decreased to 0.75 at iteration 58. The pretence attack on MNIST resulted in an accuracy of 0.89 at iteration 58 and the randomized attack has an accuracy of 0.15 at the epoch. In this use case, we notice that the attack that has the biggest impact is the randomized attack where any \textit{worker} can be compromised with a certain probability by the adversary at every iteration.

When the defense is applied to protect against the static, pretence and randomized attack, we notice in all the figures that the accuracy of the model improves. When the defense starts at $\Delta$ = 0, in case of a static attack on MNIST, the model's accuracy is 0.72 at iteration 58. In Figure~\ref{fig:MNIST_2}, for the pretence attack, the accuracy improved of 0.07 at iteration 58 when $\Delta$ = 0 compared to when no defense is applied. Similarly for the randomized attack on MNIST in Figure~\ref{fig:MNIST_3}, the accuracy also improved of 0.07 at iteration 58 when $\Delta$ = 0. 

This is due to the robustness of the federated learning that is different under fixed or variable ratios of malicious \textit{workers}. The ratio can be fixed (static attack) or variable (pretence, randomized attacks). After \textit{t} steps of gradient descent when the ratio is fixed, the algorithm asymptotically converge to the global optimum. This is due to the fact that when the ratio is fixed over time, the defense helps the underlying system attain a sub-optimal parameter with a certain error in a predictable number of steps. The convergence of the learning process has the optimal rate O(1/t) in Byzantium-free learning case (baseline curves on the figures). Figure~\ref{fig:MNIST_1} shows an empirical validation of the analytical result for the MNIST dataset when the ratio of malicious \textit{workers} is fixed. However, when the ratio of malicious \textit{workers} varies over time as in Figures~\ref{fig:MNIST_2} and ~\ref{fig:MNIST_3}, the defense detects different rates from iteration to iteration so the error is not predictable, thus the accuracy is not stable over the training period.

On another hand, we notice that for the same attack, when the defense is applied at different periods of time $\Delta$=10 and $\Delta$=40, the accuracy of the model is different. In Figure~\ref{fig:MNIST_1} for the static attack, at $\Delta$=0, the accuracy reaches 0.72 at iteration 58. However, when the defense starts at $\Delta$=10, the accuracy is 0.82 and at $\Delta$=40, the accuracy is 0.78. For this attack, we realize that between $\Delta$=0 and $\Delta$=10, the accuracy increased. This means that even if we start the defense 10 iterations later, it will not have a bad impact but it would rather improve the accuracy of the model. For the pretence attack, between $\Delta$=0 and $\Delta$=10, the accuracy increased of 0.01 and for the randomized attack, it increased of of 0.14. This shows that the best protection against attacks does not need to start right at the beginning of training but can start later and even performs better. There is an optimum amount of time where if the defense starts, the federated learning process can still be protected against attacks in an efficient way. We call this period of time, a monitoring period of duration $\Delta$.

However, we notice that for all attacks, if the defense was to be applied later, at $\Delta$=40, the accuracy decreases significantly compared to when it is applied at $\Delta$=10. For the static attack on MNIST, the accuracy decreased of 4\% for the defense at $\Delta$=40 compared to when it was started at $\Delta$=10. Similarly for the pretence attack and the randomized attack on MNIST where the accuracy decreased of 7\% and 14\% respectively between $\Delta$=10 and $\Delta$=40 at iteration 58 in Figures~\ref{fig:MNIST_2}, ~\ref{fig:MNIST_3}. This indicates that the defense must not start too late otherwise the training is impacted by the attacks. This result was expected because the attackers had lots of time to sabotage the system. The defense that came later was not able to increase the model's accuracy in comparison to the defense that was applied earlier and was already eliminating malicious nodes from the FL process. We notice that for this use case, the randomized attack is the most impacted by a delay before a defense is applied.

\begin{figure}[h]
\begin{center}
  \includegraphics[width=1\linewidth]{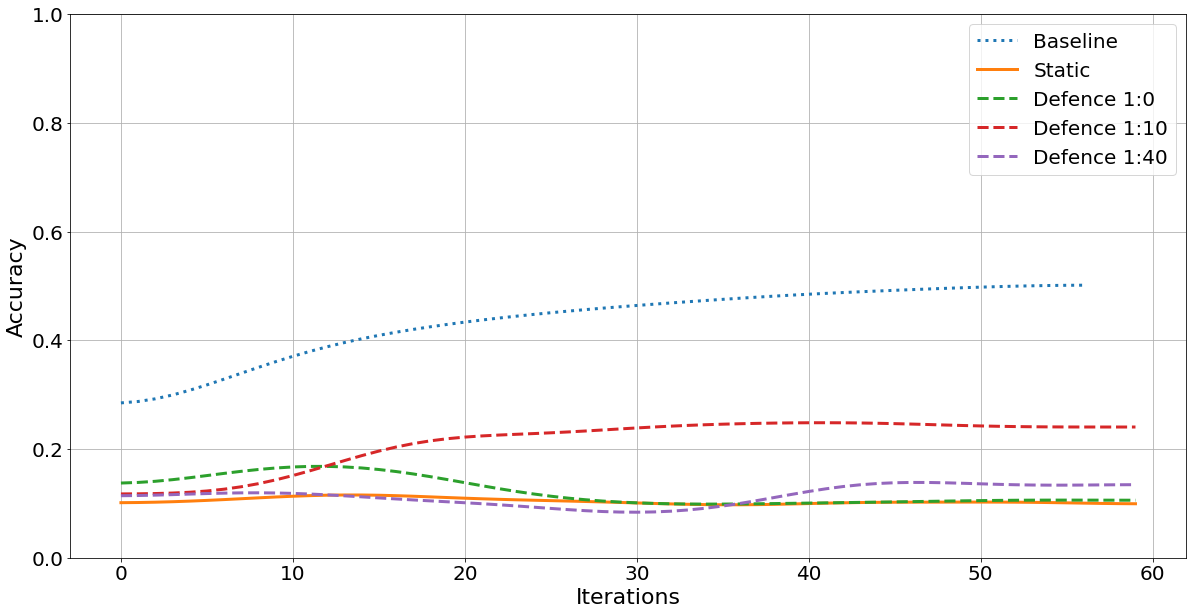}
  \caption{CIFAR - Accuracy of the global model when under a static attack with the defense starting at different $\Delta$ periods.}
  \label{fig:CIFAR_1}
\end{center}
\end{figure}

\begin{figure}[h]
\begin{center}
  \includegraphics[width=1\linewidth]{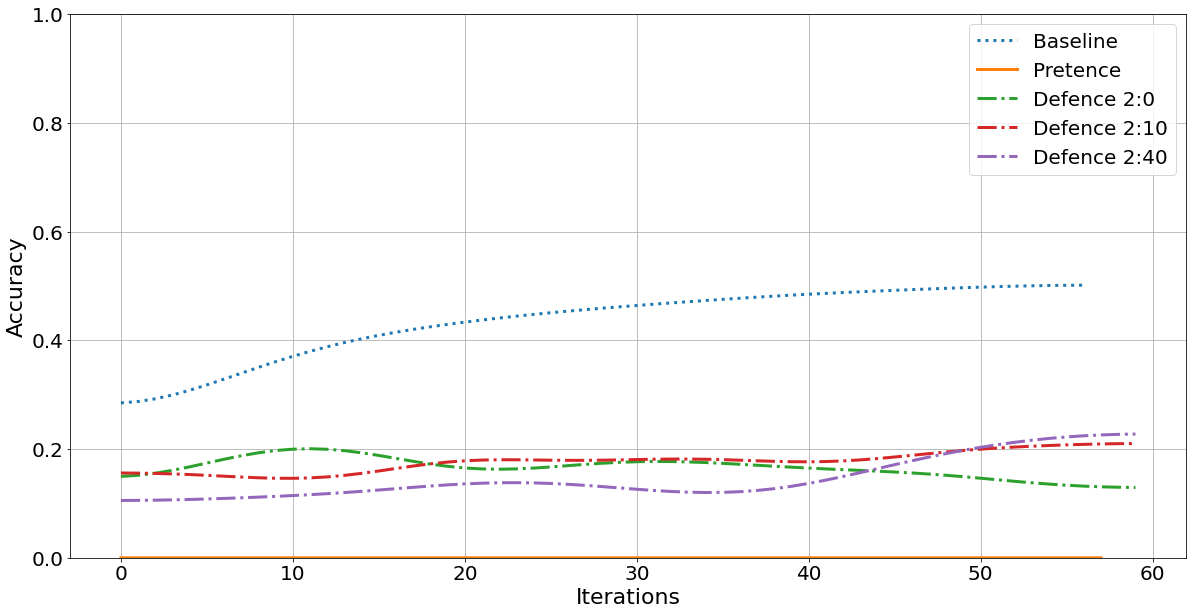}
  \caption{CIFAR - Accuracy of the global model when under a pretence attack with the defense starting at different $\Delta$ periods.}
  \label{fig:CIFAR_2}
\end{center}
\end{figure}

\begin{figure}[h]
\begin{center}
  \includegraphics[width=1\linewidth]{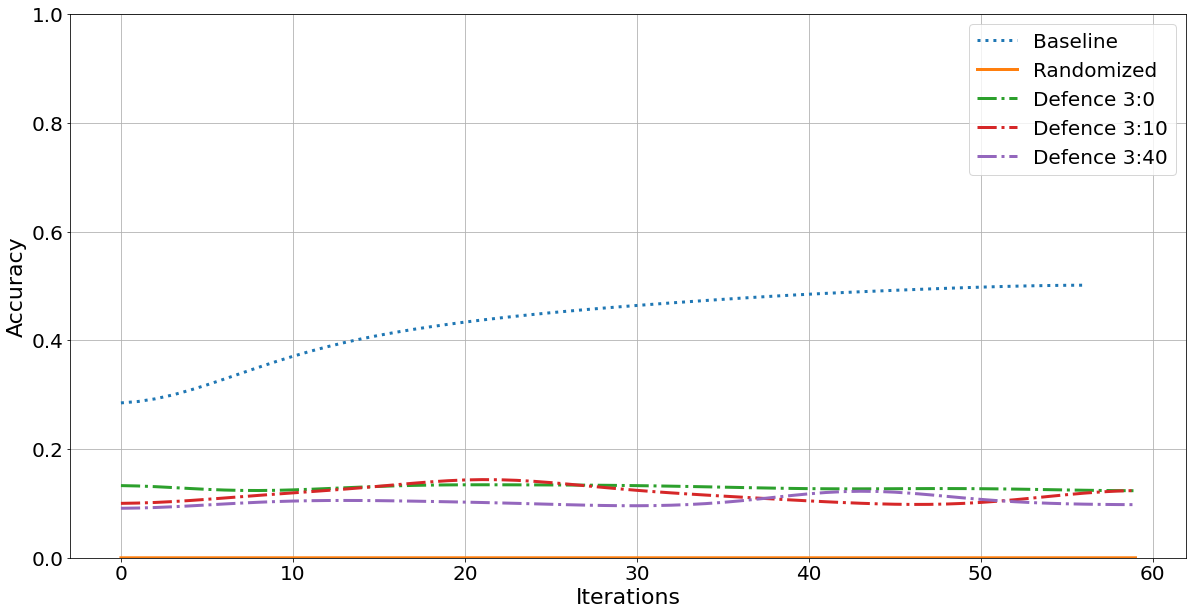}
  \caption{CIFAR - Accuracy of the global model when under a randomized attack with the defense starting at different $\Delta$ periods.}
  \label{fig:CIFAR_3}
\end{center}
\end{figure}

Figures~\ref{fig:CIFAR_1}, ~\ref{fig:CIFAR_2}, and  ~\ref{fig:CIFAR_3} show the accuracy of the global model for the CIFAR-10 use case under the static, pretence and randomized attacks with the defense starting at different $\Delta$ periods. When the defense is applied to protect against the attacks, we notice in all the figures that the accuracy of the model improves.

Similar to MNIST, when the defense starts at $\Delta$=10 in Figure~\ref{fig:CIFAR_1}, the model's accuracy increases of 0.13 at iteration 58 in comparison to when the defense starts at $\Delta$=0. This means that even if we start the defense 10 iterations later, it will not have a bad impact but it would rather improve the accuracy of the model. In Figure~\ref{fig:CIFAR_2}, for the pretence attack on CIFAR, the accuracy improved of 0.08 at iteration 58 when the defense started at $\Delta$ = 10 compared to when it started at $\Delta$=0. For the randomized attack in Figure~\ref{fig:CIFAR_3}, the accuracy remained the same at iteration 58. 

Also, we found a similar result for CIFAR as for MNIST regarding the case when the defense is applied later, at $\Delta$=40. The defense must not start too late because the accuracy decreases compared to when it is applied at $\Delta$=10. For the static attack on CIFAR in Figure~\ref{fig:CIFAR_1}, the accuracy decreased of 10\% for the defense at $\Delta$=40 compared to when it was started at $\Delta$=10. For the pretence attack in Figure~\ref{fig:MNIST_2}, the accuracy increased of 2\%. For the randomized attack in Figure~\ref{fig:MNIST_3}, the accuracy decreased of 3\% between $\Delta$=10 and $\Delta$=40 at iteration 58. For this use case, the static attack is the most impacted by a delay before a defense is applied.

During a randomized attack on CIFAR where the accuracy remained the same, this is still an important result in terms of  improvement of system efficiency as related to time cost and computation overheads. The conclusion from the results obtained from the evaluation of the two datasets is that anomaly detection in federated learning can be delayed while the same or even better performance in terms of accuracy can be achieved. On the other hand, in terms of system performance, the monitoring period will undoubtedly decrease computation overhead and time cost. 

\begin{figure*}[!ht]
\begin{center}
  \includegraphics[width=0.73\linewidth]{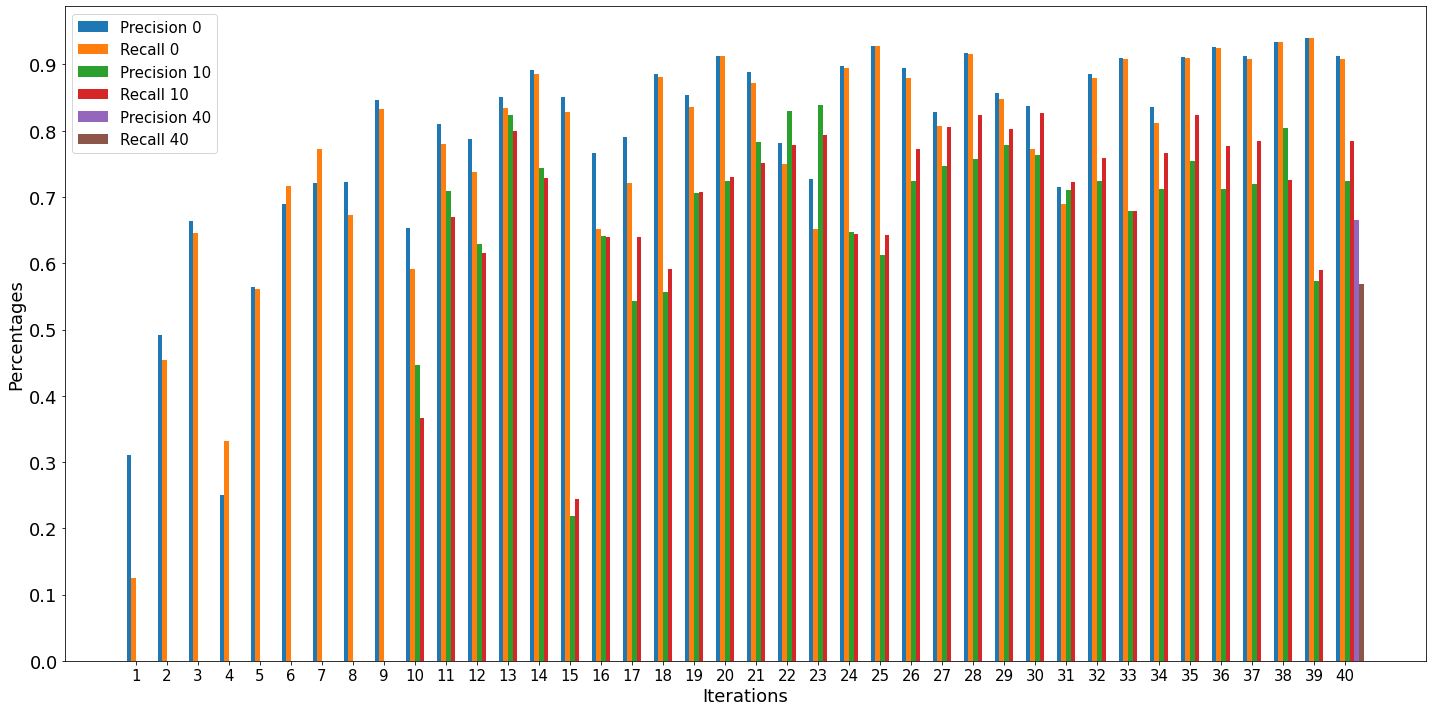}
  \caption{MNIST - Precision and recall of the defense applied at different $\Delta$ periods of time against a static attack for the first 40 iterations.}
  \label{fig:MNIST_precisionrecall1}
\end{center}
\end{figure*}

\begin{figure*}[!h]
\begin{center}
  \includegraphics[width=0.73\linewidth]{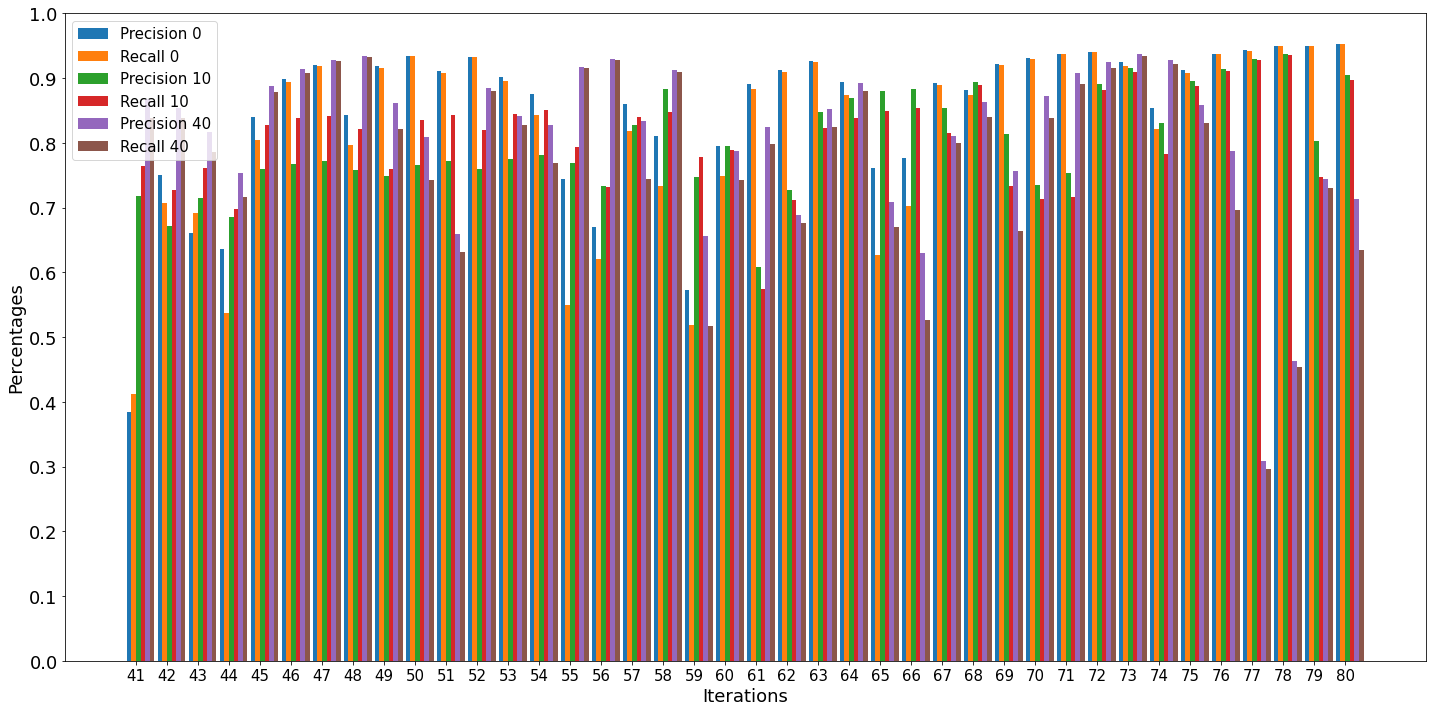}
  \caption{MNIST - Precision and recall of the defense applied at different $\Delta$ periods of time against a static attack between iterations 40-80.}
  \label{fig:MNIST_precisionrecall2}
\end{center}
\end{figure*}

To understand the reason behind the variation in the accuracy of the global model when the defense starts at different periods of time, we must look into the precision and recall scores of the detection mechanism during the first iterations of training. This will also enable us to validate the need of a monitoring period when defending the FL process against attacks. In Figures~\ref{fig:MNIST_precisionrecall1} and \ref{fig:MNIST_precisionrecall2}, we show the precision and recall in terms of percentage and the values are reported for the defense against the static attack on MNIST applied at different $\Delta$ periods of time. We do not show the figures of the similar analysis for the pretence and randomized attacks because they show a similar behavior.

When $\Delta$ = 0, we notice from Figure~\ref{fig:MNIST_precisionrecall1} that the value of precision is almost always higher than recall at every iteration. However, when $\Delta$ = 10, which corresponded in Figure~\ref{fig:MNIST_1} to the monitoring period enabling the best global model accuracy when protecting against a static attack, the values of recall are higher than precision at almost every iteration. When $\Delta$ = 40, we notice from Figure~\ref{fig:MNIST_precisionrecall2} that the value of precision is almost always higher than recall at every iteration. Thus, this does not represent a good indication that this must be the appropriate value of $\Delta$. This validates the fact that in the context of FL, a detection technique that ensures high recall but reduces precision is better than one that increases precision but reduces recall. The appropriate monitoring period can be found by looking at the values of recall and precision for a subset of \textit{workers} of the FL process. Afterwards, this period can be implemented on the the rest of the nodes of the network in order to defend against attacks in an efficient way.

\begin{figure*}[!h]
\begin{center}
  \includegraphics[width=0.73\linewidth]{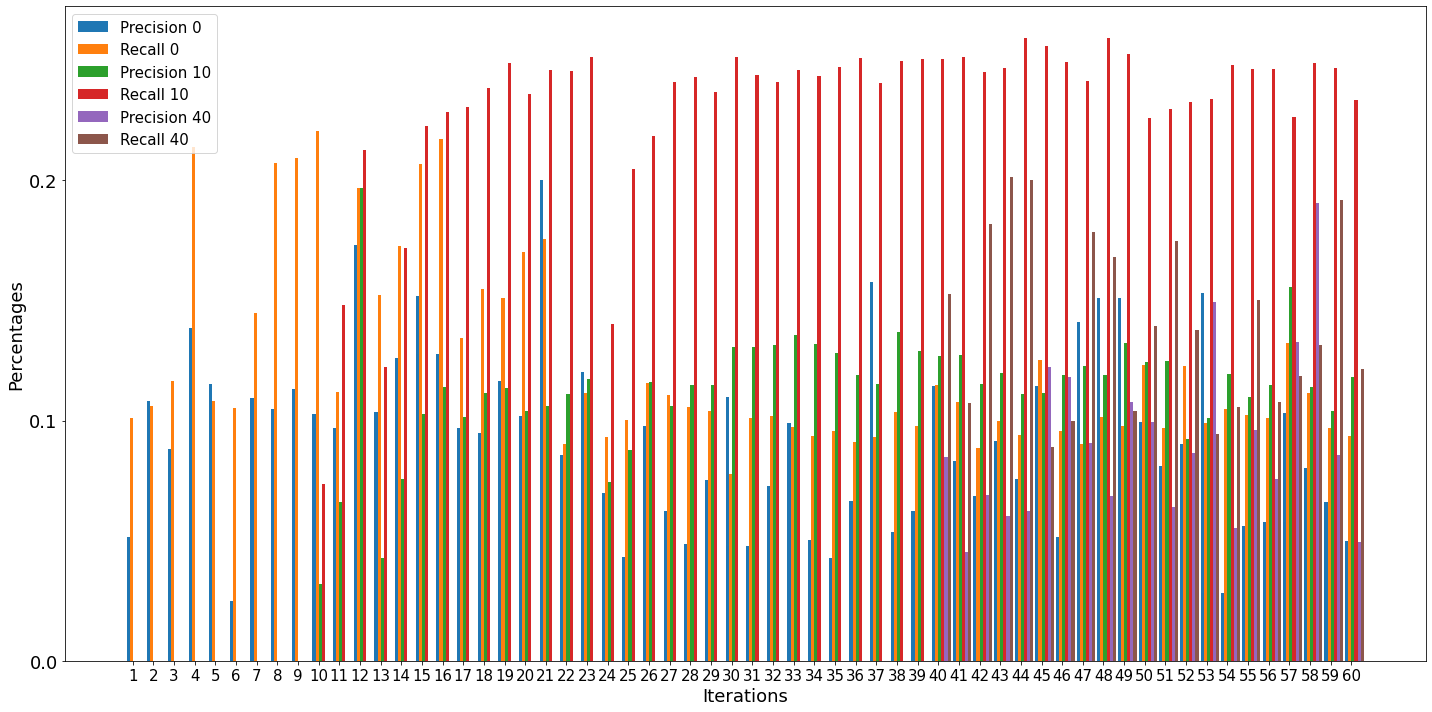}
  \caption{CIFAR - Precision and recall of the defense applied at different $\Delta$ periods of time against a static attack.}
  \label{fig:CIFAR_precisionrecall}
\end{center}
\end{figure*}

Similarly, Figure~\ref{fig:CIFAR_precisionrecall} shows the precision and recall in terms of percentage for the defense against the static attack on CIFAR applied at different $\Delta$ periods of time. The monitoring period that ensures high recall but reduces precision during training is considered the best. From the Figure, $\Delta$=10 can be distinguished by the high values of recall compared to the precision throughout the training interval considered. 

\begin{figure}[!h]
\begin{center}
  \includegraphics[width=1\linewidth]{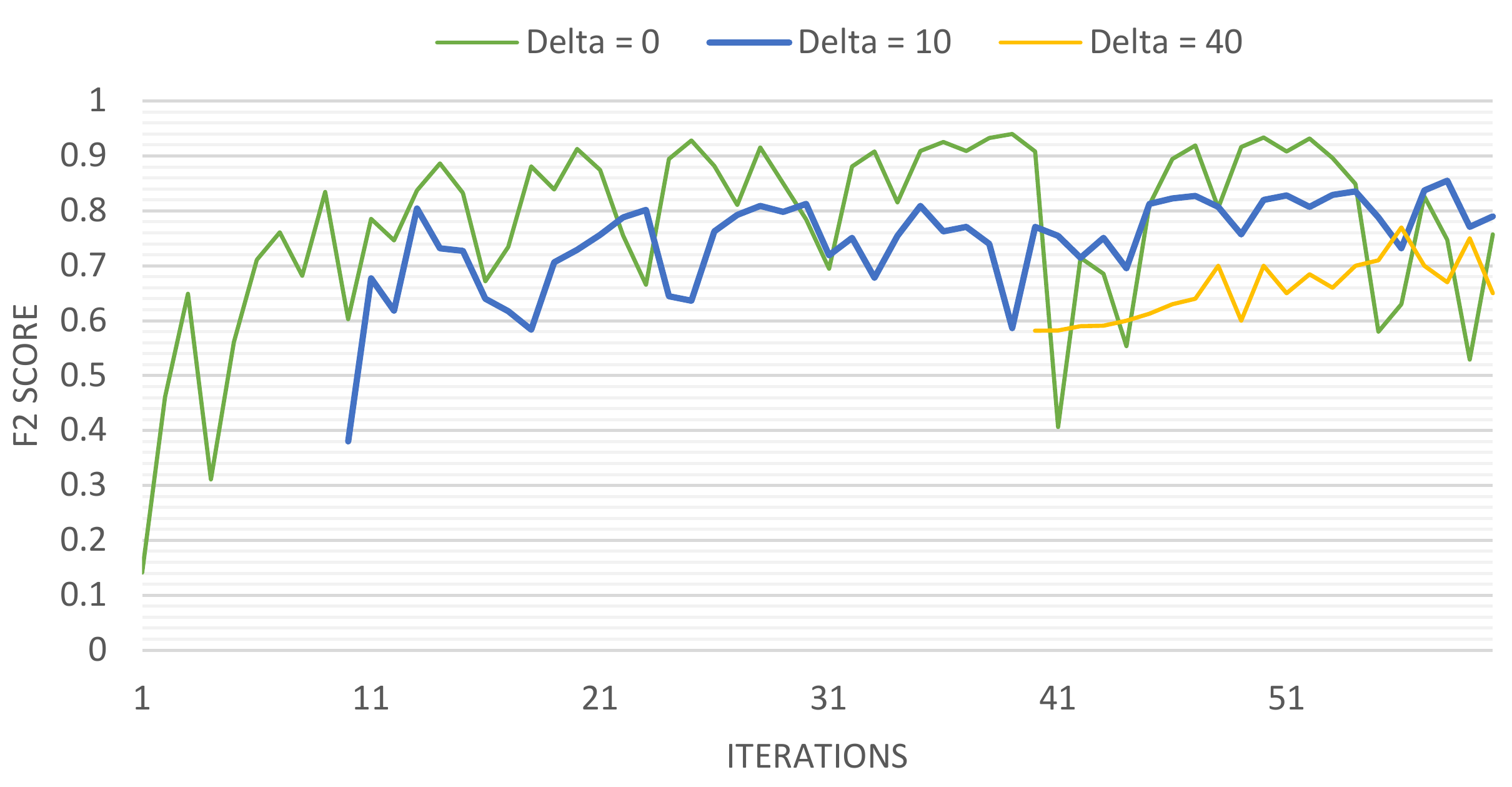}
  \caption{MNIST - $F_2$ values as a measure of the performance of the detection for different $\Delta$ periods.}
  \label{fig:MNIST_F_beta}
\end{center}
\end{figure}

\begin{figure}[!h]
\begin{center}
  \includegraphics[width=1\linewidth]{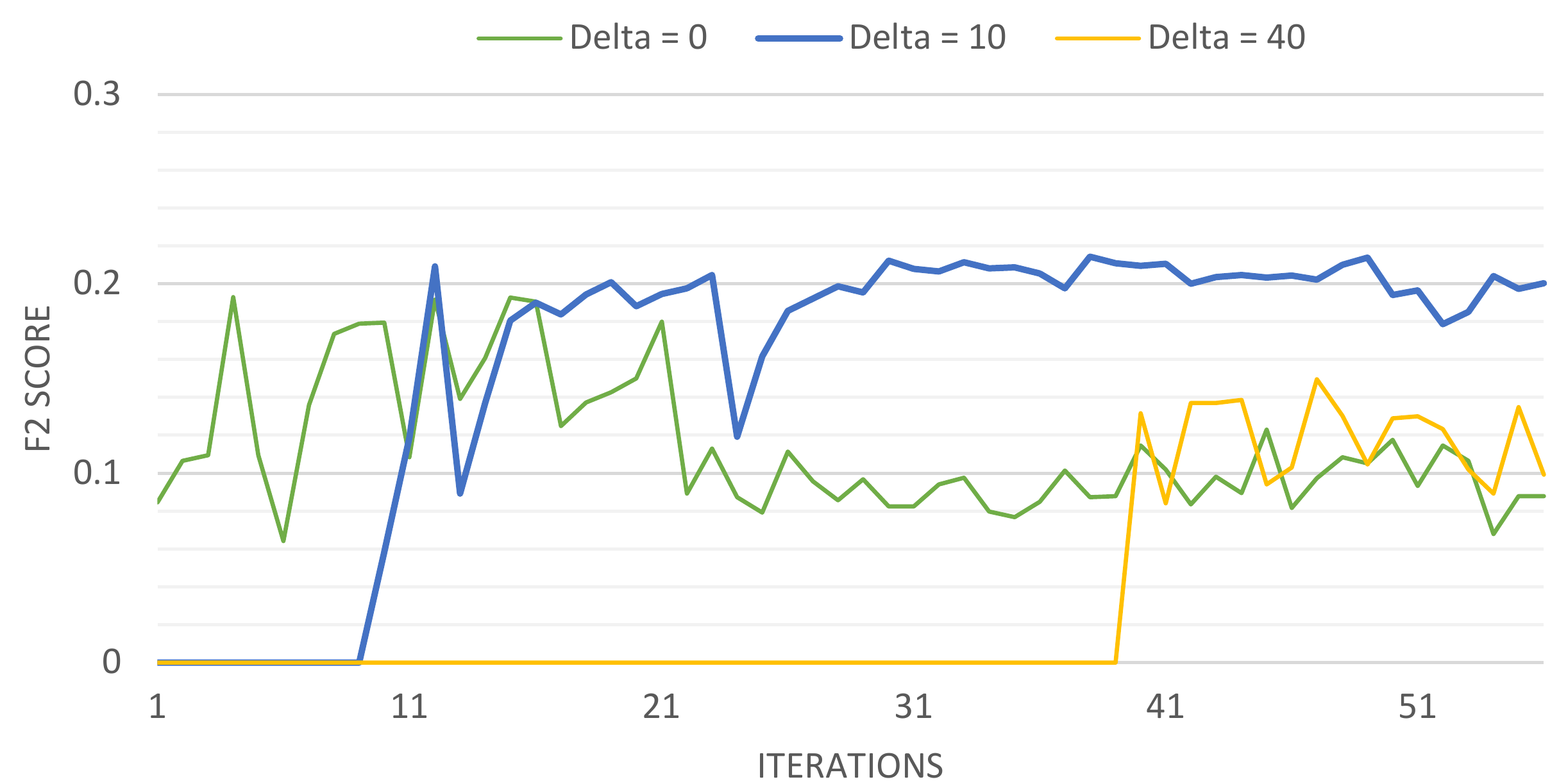}
  \caption{CIFAR - $F_2$ values as a measure of the performance of the detection for different $\Delta$ periods.}
  \label{fig:CIFAR_F_beta}
\end{center}
\end{figure}

Precisely, estimating the appropriate value of $\Delta$ must consider both precision and recall scores. We present in Figure~\ref{fig:MNIST_F_beta}, the $F_2$ values at every iteration as a measure of the performance of the defense for different $\Delta$ periods. The $F_2$ measure weights recall higher than precision. Similarly, Figure~\ref{fig:CIFAR_F_beta} presents the $F_2$ values for the CIFAR use case and they measure the effectiveness of detection with respect to a \textit{chief} who attaches more importance to recall than precision. In both use cases, low values of F-score are associated to $\Delta$ =40. This monitoring period is thus not the most appropriate for the models under study. In CIFAR, the highest values of F-score are associated to a monitoring period of $\Delta$ =10. In fact, this monitoring period is the most efficient for the CIFAR FL process. For MNIST, F-score values of $\Delta$ =10 are very close to the values attained when $\Delta$ =0, so even if the defense starts at a later time, the scores are not impacted a lot by this delay in protection against attacks. 



Finally, there is an inverse relationship between the monitoring period and the accuracy, where it is possible to increase one at the cost of reducing the other. However, as the results imply at $\Delta$=10, compared to when $\Delta$=0 and $\Delta$=40, adapting the monitoring period will bring time cost and consequently computation overheads down for the defense for both the MNIST and CIFAR use cases. 

\newpage
\section{Conclusion}
\label{sec:conclusion}

In this paper, we studied how to optimize the timing of deploying the defense mechanisms against poisoning attacks in federated learning. In fact, we proposed that a monitoring period must be accounted for particularly in defense approaches that are based on behavioral pattern analysis to eliminate potential malicious \textit{workers} from the FL process. We showed it by conducting a comparative analysis of training traces of benign and malicious \textit{workers} during the early stages of training. We evaluated the monitoring period required in order to improve system efficiency in terms of the time cost and computation overheads in the beginning of training for different benchmark systems for text and image classification (MNIST, CIFAR-10). Empirical results suggest in most cases, the correction lowers the false positives and false negatives and helps the distributed learning system achieve better performance in the early stage of training. To fine-tune the appropriate monitoring period throughout the distributed learning process, game theoretic research holds promise in addressing this challenge.

\bibliographystyle{elsarticle-num}
\bibliographystyle{IEEEtran}
\bibliography{bibtexarticle.bib}

\EOD

\end{document}